\documentclass[10pt,twocolumn,letterpaper]{article}

\usepackage[pagenumbers]{cvpr} % To force page numbers, e.g. for an arXiv version

\usepackage{graphicx}
\usepackage{amsmath}
\usepackage{amssymb}
\usepackage{booktabs}
\usepackage{multirow}
\usepackage{float}
\usepackage{gensymb}

\usepackage[pagebackref,breaklinks,colorlinks]{hyperref}

\usepackage[capitalize]{cleveref}
\crefname{section}{Sec.}{Secs.}
\Crefname{section}{Section}{Sections}
\Crefname{table}{Table}{Tables}
\crefname{table}{Tab.}{Tabs.}

\def\cvprPaperID{1544} % *** Enter the CVPR Paper ID here
\def\confName{CVPR}
\def\confYear{2023}

\begin{document}

% \title{CrossFusion: Toward Strong Noise-resistant LiDAR-Camera Fusion for Robust 3D Object Detection with Cross-modal Complementation}
% \title{CrossFusion: Exploiting cross-modal complementation for noise-robust 3D object detection}
\title{CrossFusion: Interleaving Cross-modal Complementation for Noise-resistant 3D Object Detection}

\author{Yang Yang\\
Zhejiang University of Technology\\
{\tt\small yangyang98@zjut.edu.cn}
\and
Weijie Ma\\
The Chinese University of Hong Kong(Shenzhen)\\
{\tt\small weijiema@link.cuhk.edu.cn}
\and
Hao Chen\\
Zhejiang University\\
{\tt\small haochen.cad@zju.edu.cn}
\and
Linlin Ou\\
Zhejiang University of Technology\\
{\tt\small linlinou@zjut.edu.cn}
\and
Xinyi Yu\\
Zhejiang University of Technology\\
{\tt\small yuxy@zjut.edu.cn}
}
\maketitle

%%%%%%%%% ABSTRACT
\begin{abstract}
% WJ
% The combination of LiDAR and camera modalities is proven to be necessary and typical for 3D object detection according to recent studies. Existing fusion strategies tend to overly rely on the LiDAR modal in essence, which exploits insufficiently the abundant semantics from the camera sensor. However, in complex and various outdoor conditions, LiDAR usually produces low-quality information thus severely limiting the model's performance, which exactly requires the supplementary feature from the camera to improve. Following this, we propose CrossFusion, a more robust and noise-resistant scheme that makes full use of the camera and LiDAR features with the designed cross-modal complementation strategy. Extensive experiments we conducted show that our method not only outperforms the state-of-the-art methods under the setting without introducing an extra depth estimation network but also demonstrates our model's noise resistance without re-training for the specific malfunction scenarios by increasing 9.8 mAP and 28.8 NDS.

The combination of LiDAR and camera modalities is proven to be necessary and typical for 3D object detection according to recent studies. Existing fusion strategies tend to overly rely on the LiDAR modal in essence, which exploits the abundant semantics from the camera sensor insufficiently. However, existing methods cannot rely on information from other modalities because the corruption of LiDAR features results in a large domain gap. Following this, we propose CrossFusion, a more robust and noise-resistant scheme that makes full use of the camera and LiDAR features with the designed cross-modal complementation strategy. Extensive experiments we conducted show that our method not only outperforms the state-of-the-art methods under the setting without introducing an extra depth estimation network but also demonstrates our model's noise resistance without re-training for the specific malfunction scenarios by increasing 5.2\% mAP and 2.4\% NDS.

\end{abstract}

%%%%%%%,%% BODY TEXT
\section{Introduction} \label{sec:intro}

3D object detection plays a crucial role in autonomous driving, and the quality of the detection can directly determine the safety level in practical application scenarios. Thanks to the accurate 3D information of LiDAR, some previous works \cite{zhou2018voxelnet,yan2018second,chen2022scaling} have achieved satisfactory results on the nuScenes dataset \cite{caesar2020nuscenes}. Recently, some LiDAR-Camera fusion methods \cite{bai2022transfusion, Ku2018Joint3P, Chen2017Multiview3O,chen2022futr3d} have shown more powerful perception capabilities based on the strong geometric prior knowledge of the point cloud. However, in autonomous driving scenarios, where LiDAR signals are often affected by weather, light, reflections, etc. The performance of these methods is not ideal. Due to design limitations, image information cannot amend in this situation as expected. %due to structural limitations in their inability to use image information to counteract LiDAR noise.

\begin{figure}[ht]
  \centering
    \includegraphics[width=0.9\linewidth]{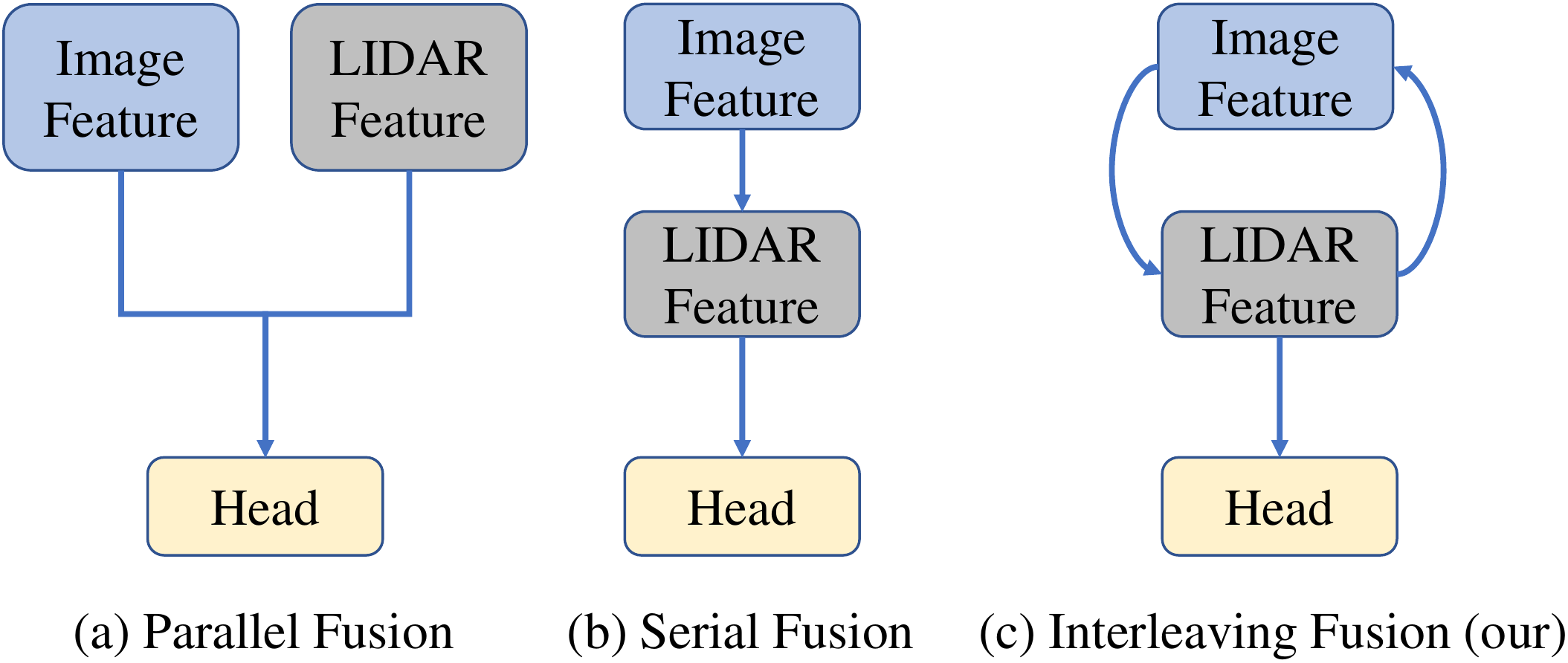}
   \caption{Comparison of LiDAR-camera fusion strategies. (a) Parallel fusion: Each branch can work independently, but the 2D-3D projection process is complex. (b) Serial Fusion: Overly rely on the LiDAR. (c) Our method: Complementary modal information through modal interleaving. }
   \label{fig:intro}
\end{figure}

As shown in \cref{fig:intro}, the existing fusion models can be mainly divided into two forms, parallel fusion and serial fusion. Parallel fusion maps image features and point cloud features into a unified space and uses the fusion features to locate the object\cite{liang2022bevfusion,chen2022futr3d,li2022uvtr}. The serial fusion, people usually use the image features from specified regions to enhance the point cloud information results\cite{Vora2020PointPaintingSF,xu2021fusionpainting,wang2021pointaugmenting},  or inject image features and point cloud features into proposal\cite{bai2022transfusion, Ku2018Joint3P, Chen2017Multiview3O}.

% In autonomous driving scenarios, LiDAR is usually affected by weather, light, and large vehicle occlusion. The instance-level fusion method, represented by Transfusion\cite{bai2022transfusion}, can experience dramatic performance degradation due to the input of low-quality LiDAR features. This situation can pose a safety hazard to autonomous driving decisions. In addition, they do not have sufficient modal interaction and therefore cannot sufficiently mitigate the gap between instance sparse point cloud features and dense image features. Although the BEV-level fusion method can alleviate the influence of LiDAR feature degradation by introducing a depth estimation network\cite{liang2022bevfusion}, the conversion process from camera to BEV is very complicated and time-consuming. Therefore, better hardware devices are needed to meet the deployment requirements.
Although these methods have achieved impressive results, they still have shortcomings. The serial fusion method has two potential problems. Firstly, their performance is inherently dependent on the precise 3D information provided by LiDAR, but they do not deeply explore the importance of images for localization. If low-quality LiDAR features are input, they may produce inaccurate results. This situation could pose a safety hazard to autonomous driving decisions. Secondly, they do not seem to adequately consider the information interaction between LiDAR and images, which may hinder the complementary properties between the modalities. Therefore, the association between geometric and depth information specific to sparse point cloud features and dense semantic features of images may be limited. 

We believe that the LiDAR-camera fusion framework's core challenge lies in exploiting complementarity. In other words, when the features provided by one sensor are contaminated, the fusion model can adaptively invoke features from the other modality to resist the perturbation of the noise and balance the weight of the two modalities.

\begin{figure*}[t]
  \centering
    \includegraphics[width=0.98\linewidth]{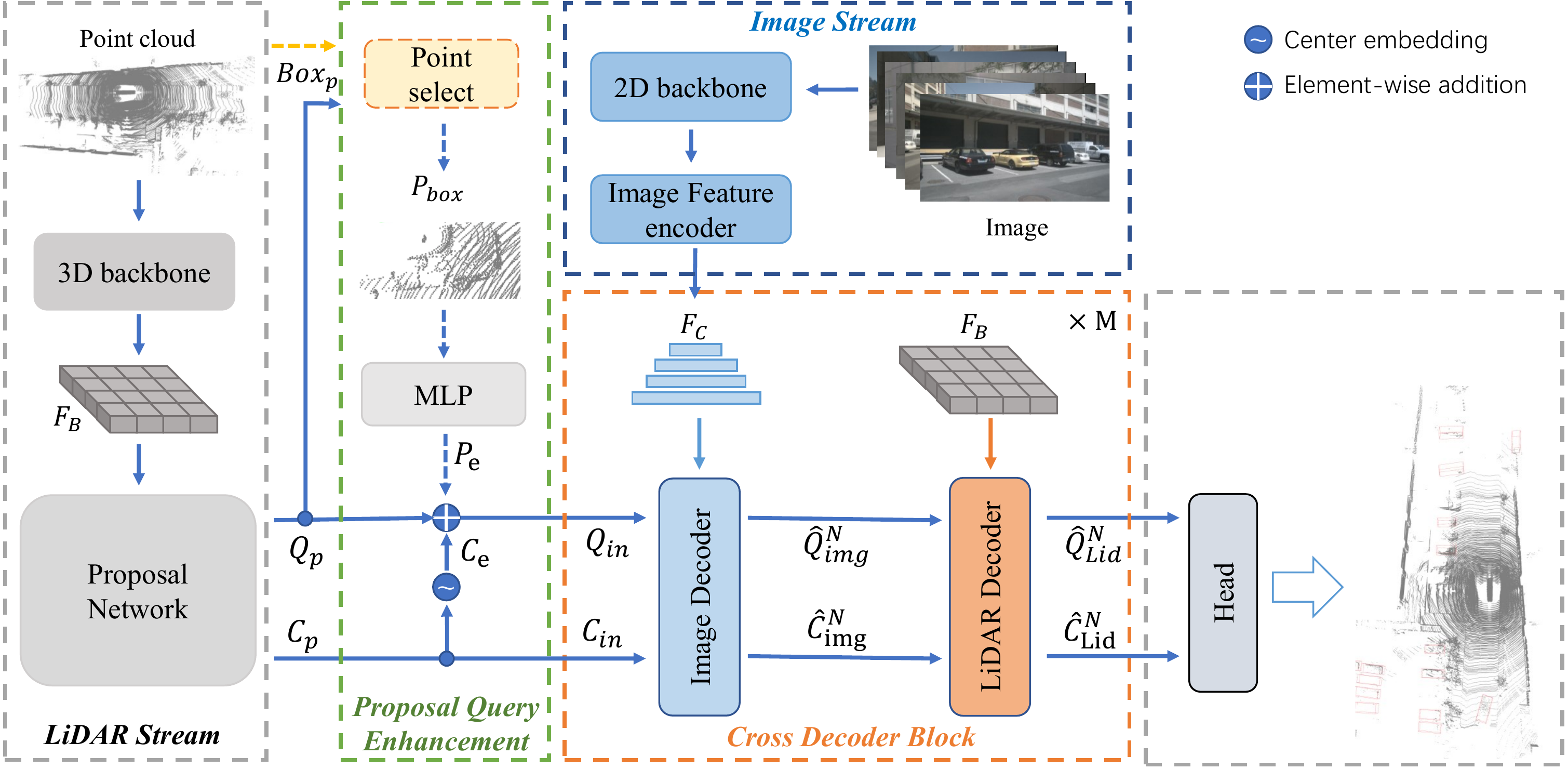}

   \caption{Overall pipeline of CrossFusion. Our model includes two main stages, and the first stage consists of the commonly used 3D backbone and proposal network, which is used to generate a set of proposal queries and the corresponding proposal centers. The second stage introduces a cross-decoder that will alternatively fuse multi-scale image features and BEV features to fine-tune the result of the proposal in the first stage.}
   \label{fig:framework}
\end{figure*}

In this paper, we design a framework for modal interleaving fusion, dubbed CrossFusion(see \cref{fig:framework}). Specifically, we design a two-stage network. First, we generate a set of proposal bounding boxes by LiDAR features, and then we enhance query features with points inside the proposal to force it to explore features of difficult samples. Next, we use cross decoder to alternately fuse image features and LiDAR features to adaptively build up the modal complementation capability to obtain better robustness. On the basis of the LiDAR decoder, we extract image features from the image feature encoder to alleviate the domain gap between the two modalities. Our contributions are summarized as follows:

\begin{itemize}
% \item Our study explores the potential limitations of the current LiDAR-camera fusion framework, namely the lack of complementarity between modalities.
\item We propose a novel LiDAR-camera fusion model for 3D target detection. It achieves modal complementarity through an interleaving fusion architecture. It also exhibits excellent robustness against various LiDAR information quality degradations without re-training.
\item We introduce a pluggable feature enhancement operation to the proposal query, which encourages the network to learn hard samples and balances the contribution of image features and LiDAR features to perception.
\item We achieve state-of-the-art methods under the setting without introducing an extra depth estimation network.
\end{itemize}

\section{Related Works} \label{sec:related}

Below are two main categories of 3D detection methods based on the amount and type of modality.

\noindent \textbf{Single-sensor}. There are two main types of single-sensor 3D detection, which are LiDAR-only and camera-only. Point clouds are extensively used in 3D detection as a fundamental and effective data source in autonomous driving. One of the common ways is to directly operate on the raw LiDAR point clouds \cite{qi2017pointnet, qi2017pointnet++, shi2019pointrcnn, yang20203dssd, yang2019std, qi2018frustum}. 
And another is based on transforming the point clouds onto a regular space, such as 3D voxels\cite{zhou2018voxelnet}, range views \cite{sun2021rsn, fan2021rangedet}, or pillars \cite{wang2020pillar, lang2019pointpillars, yin2021center}. 
However, the measurable range of LiDAR is relatively narrow and sparse, limiting the observation of distant objects. In contrast, camera images possess a greater range and relatively dense information. This has led to the emergence of camera-only detection methods. Analogous to 2D detection methods, early methods usually predict the 3D bounding boxes based on the 2D bounding boxes \cite{brazil2019m3d, mousavian20173d, xu2018multi, simonelli2019disentangling, zhou2019objects}. Recent works further develop the direct way based on the 2D advanced detectors. FCOS3D \cite{wang2021fcos3d} adapts a 2D-guided multi-level 3D detection and  directly predicts 3D bounding boxes for each object based on FCOS \cite{tian2019fcos}. Following Deformable-DETR \cite{zhu2020deformable}, DETR3D \cite{wang2022detr3d} samples the corresponding features by warping learnable 3D queries in 2D images for end-to-end 3D bounding box prediction without NMS post-processing. Since the camera view is like frustum, which isn't conducive to observation in the world coordinate system, various methods \cite{wang2019pseudo, reading2021categorical} use a bird's-eye view (BEV) 2D space transformation to unify the construction of information from the image in each camera view. Inspired by Lift-Splat-Shoot (LSS) \cite{philion2020lift}, many methods \cite{huang2021bevdet, xie2022m} further refine the transform referring to the supervision on improved depth estimation.

\noindent \textbf{Multi-sensor}. Both LiDAR and camera have their own advantages as mentioned above, so people started to explore the unification of the two modalities, and a lot of work has proven this to be reasonable and powerful. 
We categorize the existing fusion method into several forms, parallel fusion, serial fusion, and others. The Parallel fusion methods represent the work of  BEVfusion \cite{liang2022bevfusion}, UVTR \cite{li2022uvtr}, which feeds image features into the depth estimation network to obtain pseudo-point cloud features and concatenate them with point cloud features to obtain BEV and fed into the detection head to localize the object. 
As for serial fusion, there is usually a semantic prior used to finding the basic association between LiDAR and image in advance. PointPainting\cite{Vora2020PointPaintingSF} adopts the result of semantic segmentation as the prior as well as FusionPainting \cite{xu2021fusionpainting}. The proposals of bounding boxes have also been utilized by early methods \cite{Ku2018Joint3P, Chen2017Multiview3O}. However, because calibration matrices have formed a rigid link between points and pixels, such approaches are easily impacted by sensor misalignment. TransFusion \cite{bai2022transfusion} proposed a more effective proposal-based pipeline instead of the previous simple point-wise concatenation, enriching the contextual relationships between two modalities. Others like FUTR3D \cite{chen2022futr3d}, achieve an end-to-end multi-modality 3D prediction by directly fusing all modality features following DETR3D \cite{wang2022detr3d}.
% As for the serial fusion, these instances can be represented by semantic segmentation results\cite{Vora2020PointPaintingSF,xu2021fusionpainting}, using a semantic prior to finding the association between LiDAR and image, by proposals\cite{bai2022transfusion, Ku2018Joint3P, Chen2017Multiview3O} to inject image features and point cloud features into proposals, or by query\cite{chen2022futr3d}, all modal features are adaptively fused in it.
In our work, we explore a novel multi-modal fusion strategy to improve the robustness and performance of 3D detection from the perspective of mutual complementation between modalities.
% single-sensor-based approaches suffer from the vital performance bottlenecks in spite of the proliferating related works.

\section{Methods} \label{sec:method}
% WJ
In this section, we present our proposed method CrossFusion for 3D object detection. As shown in \cref{fig:framework}, given the LiDAR point clouds and the corresponding multi-view camera images as inputs, CrossFusion first extracts the LiDAR BEV features $F_L$ and multi-scale camera feature $F_C$ from their backbones in two modal streams, respectively. Then the proposal network generates the proposals of bounding boxes and selects point clouds in the box proposals for proposal query enhancement. Finally, we introduce the cross-decoder blocks to refine the bounding boxes of objects by alternately aggregating the encoded features from the modality bridge encoder and LiDAR BEV features. Each component will be introduced in the following subsections in detail.

% In this section, we present the proposed method CrossFusion for noise-resistant multi-modal fusion for 3D object detection, as well as the design motivation of the fusion approach. As shown in \cref{fig:framework}, given a LiDAR point cloud and synchronized multi-view camera images as inputs, CrossFusion first extracts multi-scale image features and BEV(Bird's-eye view) features from backbones. Then proposal network generates proposal bounding boxes and selects the point cloud from the proposal boxes $Box_{p}$.
% Finally, cross-decoder modules are used to refine the bounding boxes by aggregating the multi-scale image features from the image feature encoder and BEV features. Each component will be introduced in the following subsections.

\subsection{Image Feature Encoder}
% WJ
Despite acquiring the high-level features of two modalities from their respective backbone, there is a certain gap between the image features and LiDAR BEV features. Different from the previous LiDAR-Camera fusion strategy that directly fuses two features \cite{chen2022futr3d,bai2022transfusion}, we establish an image-view feature encoder before passing the image feature into the subsequent fusion to enrich the image semantic in the common space and bridge the modal gap. 

Specifically, we use the feature maps from the backbone in the image stream with resolutions $\tfrac{1}{32}$, $\tfrac{1}{16}$, and $\tfrac{1}{8}$ of the size of the original images. Following \cite{zhu2020deformable}, we add positional embedding and scale-level embedding to the multi-scale feature and adopt the deformable attention to enhance the feature further. We denote the output as $F_{C}^{l} \in \mathbb{R}^{d_{C}\times H_{C}^{l} \times W_{C}^{l}}$, where $l$ is the number of feature levels and $H_{C}\times W_{C}$ is the size of feature map and $d_{C}$ means the dimension of feature.

% It is well known that there is a certain difference between image features and high-level features of LiDAR. Different from the previous LiDAR-Camera fusion strategy that directly fuses image and LiDAR high-level features \cite{chen2022futr3d,bai2022transfusion}, we propose to add an Image Feature Encoder to reduce the gap.

% We use the feature maps with resolutions 1/32, 1/16, and 1/8 of the original images from an image backbone. Following \cite{zhu2020deformable} we add positional embedding and scale-level embedding to each resolution feature. It outputs multi-scale features $F_{C}^{l} \in \mathbb{R}^{C_{C}\times H_{C}^{l} \times W_{C}^{l}}$, where $l$ is the number of feature levels and $H_{C}\times W_{C}$ is the shape of feature map. 

% \begin{figure*}[t]
%   \centering
%   % \fbox{\rule{0pt}{2in} \rule{0.9\linewidth}{0pt}}
%     \includegraphics[width=0.8\linewidth]{fig/crossdecoder.pdf}
%    \caption{Illustration of the Cross Decoder Block}
%    \label{fig:crossdecoder}
% \end{figure*}
\subsection{Proposal Query Enhancement}
% WJ
The query initialization in the Transformer architecture not only accelerates the training process but also improves the model accuracy, according to many excellent works for 2D detection \cite{zhu2020deformable,meng2021conditional}. Referring to this, as illustrated in \cref{fig:framework}, we design two query enhancement strategies to strengthen the representation of query for later decoding. And before that, following \cite{bai2022transfusion}, we first adopt the input-dependent initialization to get the object queries $Q_p \in \mathbb{R}^{N \times d}$ from the proposal network, where $N$ is the number of the proposal and $d$ is the dimension of the whole model.

% The query initialization in the Transformer architecture not only accelerates training but also improves model accuracy \cite{zhu2020deformable,meng2021conditional}. Therefore, as illustrated in \cref{fig:framework}, we also use two enhancements to strengthen the representation of query output from the proposal network. 

% WJ
\noindent \textbf{Point Selective Augmentation}. Although the initial queries are favorable to improve the model training, as a double-edged sword, it can also bring over-reliance on the prior as well as the reluctance to focus on hard targets. In order to enhance the sensitivity to hard targets and reduce the over-dependence on the results of proposals, we propose a query augmentation strategy only during training, named Point Select Enhancement. Its main insight is to add a learnable but controllable bias to the queries, forcing our model to learn the similarity to the proposal boxes rather than to chase the consistency, thereby improving the model's robustness. 
Specifically, given the proposal bounding boxes $Box_p$ predicted by $Q_p$, we first select $Z$ random points $P_{rand} \in \mathbb{R}^{Z \times N\times d_{pc}}$ for an interval of candidates, where $d_{pc}$ is the dimension of the point cloud. Then we can get the point selective embedding $P_{box} \in \mathbb{R}^{Z\times N\times d}$ via the bias projection $f_{bp}$, which is a Multilayer Perceptron (MLP) with the dimension of $d$. Finally, we will use max pooling to acquire the local optimal response $P_{e} \in \mathbb{R}^{N\times d}$. The formulations are as follows:
\begin{equation}
    P_{box} = f_{bp}(P_{rand})
\end{equation}
\begin{equation}
    P_e = MaxPool(P_{box})
\end{equation}

% \noindent \textbf{Point Select Enhancement}. In order for our model to have the ability to learn some hard targets in BEV features and to reduce the dependence on the results of proposals. We propose a query augmentation process used only during training, named Point Select Enhancement. Our core idea is to add a learnable bias to the query, forcing the model to learn the bounding box that is similar to the proposal box and improving the robustness of the model. Specifically gives K randomly selected points in the proposal bounding box $P_{box} \in \mathbb{R}^{K\times C_{in}}$ where $C_{in}$ is the dimension of point cloud, fed into the multilayer perceptron(MLP) and use the max operator.

\noindent \textbf{Proposal Center Embedding}. 
% WJ
Unlike the position information of the query is fixedly encoded or learnably parameterized \cite{carion2020end, zhu2020deformable}, we compensate it with the center-point prediction of the initial query. This way, the query is equipped with a prior position for better observation of the object. Specifically, we embed the 3D center point $C_{in} \in \mathbb{R}^{N \times 3}$ by a $d$-dimensional MLP and add the proposal center embedding $C_e \in \mathbb{R}^{N \times d}$ to $Q_p$.

On the whole, the input query is formulated below with the element-wise addition:
\begin{equation}
    Q_{in}^{0} = Q_p+P_e+C_e
\label{eq:query_init}
\end{equation}
where ``0'' means the original input cross decoder blocks.

% In addition, we add center embedding, since the features of the proposal bounding box are all from BEV, in order to adapt the query to the subsequent cross-modal feature form, we use MLP embedding proposal-center $C_{in}$ to implicitly construct the connection between the above two modalities.
% \begin{equation}
%     C_e = MLP(C_{in})
% \end{equation}

% Finally, the input query is computed by the element-wise addition.
% \begin{equation}
%     Q_{in}^{0} = Q_p+P_e+C_e
% \label{eq:query_init}
% \end{equation}

% Secondly, we add center embedding, since the features of the proposal bounding box are all from BEV, in order to adapt the query to the subsequent cross-modal feature form, we use MLP embedding proposal-center $C_{in}$ to implicitly construct the connection between the above two modalities.
% \begin{equation}
%     C_e = MLP(C_{in})
% \end{equation}

% Finally, the input query is computed by the element-wise addition.
% \begin{equation}
%     Q_{in}^{0} = Q_p+P_e+C_e
% \label{eq:query_init}
% \end{equation}

\subsection{Cross Decoder Block}
% WJ
Different from the previous fusion approaches that combine the individual features separately \cite{bai2022transfusion} or directly merge multi-modal features into a space \cite{chen2022futr3d,liang2022bevfusion}, our core idea is to take advantage of the respective modal strengths and to update the detection results through interactive decoding. Precisely, 2D images have the advantage of high recall from a wide range, and 3D point clouds have the advantage of precision from accurate geometric coordinates. To carry forward it, we stack multiple cross-decoder blocks for interleaving cross-modal fusion. To be specific, we alternated with an image decoder and a LiDAR decoder, which enables the models to alleviate the gap between the point cloud and image features and aggregate the key features between multi-modal features adaptively. In particular, the blocks built by image-LiDAR cross-decoders will increase the recall of objects located in sparse point clouds and use LiDAR-decoder to correct for these additional targets. Below are the detailed analysis and formulation of the proposed block.

\noindent \textbf{Cross-modal Decoding Analysis}. Many works have shown the necessity of image features at the decoding stage \cite{li2022uvtr, bai2022transfusion, liang2022bevfusion}, but those works may not have focused well on the specific binding order. We discover it's some inappropriate to directly combine the image feature with the bounding box proposals previously obtained from point clouds. Concretely speaking, a single frame of the 2D image can only provide relative position information among objects and cannot acquire their absolute depths, while point clouds are capable to provide accurate 3D spatial information. This poses a problem in that the information obtained only from the camera space will lead to a relatively large bias if directly optimizing the previous proposal representations. In other words, most of the previous fundamental information underlying LiDAR space turns out to be overwhelmed. To mitigate this, we still require the LiDAR information to solidify the optimization process. 

\noindent \textbf{Image Decoder}. Attention mechanism has been widely used in 3D detection tasks as a popular way of multi-modal information interaction\cite{liang2022bevfusion, chen2022futr3d}. Here we set the image decoder layer similar to the design of Deformable DETR \cite{zhu2020deformable}. We use the set of N input queries $Q_{in} \in \mathbb{R}^{N \times d}$ and the corresponding proposal bounding box centers $C_{in} \in \mathbb{R}^{N \times 3}$. First, we project every center $(x_i, y_i, z_i)$ in $C_{in}$ into the camera coordinate system by using the calibration matrices as 2D points $(u_i,v_i)$ in projected 2D views like \cref{eq:projection}. 
\begin{equation}
\alpha\left[\begin{array}{c}
u_i\\
v_i\\
1
\end{array}\right]=\left[\begin{array}{ccc}
f_{x} & 0 & u_{0} \\
0 & f_{y} & v_{0} \\
0 & 0 & 1
\end{array}\right]
\left[\begin{array}{c}
x_i\\
y_i\\
z_i
\end{array}\right]
  \label{eq:projection}
\end{equation}
% WJ
where $f_{x},f_{y},u_{0},v_{0}$ are camera intrinsic parameters, $\alpha$ is the scale factor. After that, we regard these 2D points as the reference points of the query set $Q_{in}$ and sample the features from the projected views around these reference points. Eventually, we perform a weighted sum of the sampled features as the output of image-level cross-attention. Formally, let $F_{C}$ be the input multi-scale image feature maps and the 2D reference points ${r_{C}} \in [0,1]$ be normalized by the shape of the image feature maps. For each query, we update query in each cross decoder block $B_m$ as:
% 
% where $f_{x},f_{y},u_{0},v_{0}$ are camera intrinsic parameters, $d$ is the scale factor. Then We leverage multi-scale image feature maps for handling distant objects with sparse point clouds in the LiDAR proposal. In addition, the image decoder interacts with surrounding pixels and across multi-scale feature maps, thus enhancing the ability of the model to detect different scale targets. Formally, let $F_{C}$ be the input multi-scale image feature maps and camera reference points ${r_{c}} \in [0,1]$ be normalized by the shape of the image feature maps. For each query element $Q^{n}$, we update query in each cross decoder block $N \in \{1,2\dots,N \}$ as:
% 
\begin{equation}
    {Q}_{u}^{m}=\sum_{p=1}^{P} \mathbf{W}_{p}\sum_{l=1}^{L} \mathbf{W}_{l}\left[\sum_{k=1}^{K} A_{l k p} \mathbf{F}_{C}^{l}\left(r_{C}+\Delta x_{l k p}\right)\right]
  \label{eq:image_decoder}
\end{equation}

\begin{equation}
    \hat{Q}_{img}^{m}={Q}_{u}^{m}+Q_{img}^{m} 
  \label{eq:image_decoder_2}
\end{equation}

where $m \in \{1,2\dots,M \}$, $M$ is the number of blocks. And $k$ is the index of the sampling point, $\Delta x_{l k p}$ and $A_{l k p}$ denote the sampling offset and the attention weight of the $k$-th sampling point in $l$-th feature level and $p$-th camera, bilinear interpolation $\mathbf{F}_{C}^{l}\left(r_{C}+\Delta x_{l k p}\right)$ is applied as in computing.

\noindent \textbf{LiDAR Decoder}. The structure of this decoder resembles the image decoder above. We use the query $\hat{Q}_{img}^{m}$ and box center $\hat{C}_p^m$ output from the image decoder as the input to the LiDAR decoder. Next, we warp the box center into the BEV feature as the LiDAR-space reference point $r_{l}$ and sample the BEV Feature for updating. 
\begin{equation}
\hat{Q}_{LiD}^{m}=\sum_{k=1}^{K} A_{k} \mathbf{F}_{L}\left(r_{L}+\Delta x_{k}\right)+\hat{Q}_{img}^{m}
  \label{eq:LiDAR_decoder}
\end{equation}
where $k$ is the index of the sampling point, $\Delta x_{k}$ and $A_{k}$ denote the sampling offset and the attention weight of the $k$th sampling point, bilinear interpolation is applied as in computing $\mathbf{F}_{L}\left(r_{L}+\Delta x_{k}\right)$. It is worth noting that in the autonomous driving scenario, unlike images, targets on BEV are usually in the form of small objects, so we only selected the largest scale BEV feature maps to participate in the calculation.

% WJ
\noindent \textbf{Iterative 3D Bounding Box Refinement}. Previous work has demonstrated that the iterative refinement is an effective form of improving detection performance \cite{zhu2020deformable,wang2022detr3d}. Thus, we establish a similar iterative update mechanism between the image decoder and LiDAR decoder of every cross-decoder block. Specifically, we first send the query output from the image decoder to the regression branches $f_{reg\_img}$ and continue to apply the offset prediction to update the bounding box center in the LiDAR decoder with its branches $f_{reg\_LiD}$. The formulation is as follows:

% Previous work has demonstrated that the iterative refinement is an effective form of improving detection performance \cite{zhu2020deformable,wang2022detr3d}. Thus, we establish a similar iterative update mechanism between the image decoder and LiDAR decoder in the each cross block.
% In the LiDAR decoder, we send the query output from the image decoder to the MLP to get the offset to update the bounding box center. 
\begin{equation}
\hat{C}^{m}_{s}=C_{in}^{m}+f_{reg\_s}(\hat{Q}^{m}_{s})
  \label{eq:box_refine_img}
\end{equation}
% \begin{equation}
% \hat{C}^{m}_{LiD}=C_{in}^{m}+f_{reg\_LiD}(\hat{Q}^{m}_{LiD})
%   \label{eq:box_refine_lidar}
% \end{equation}
% where $\hat{C}_{LiD}^{m}$ is calculated by Eq.\eqref{eq:box_refine_lidar}. 
where $s$ denotes the specific modality $\in \{img, LiD\}$.
Here, each decoder block $B_m$ refines the prediction results of the previous cross-decoder block $B_{m-1}$. 
\begin{equation}
\hat{Q}_{in}^{m}=\hat{Q}_{LiD}^{m}+\hat{Q}_{in}^{m-1}
  \label{eq:block_refine_q}
\end{equation}
\begin{equation}
\hat{C}_{in}^{m}=\hat{C}_{LiD}^{m}+\hat{C}_{in}^{m-1}
  \label{eq:block_refine_c}
\end{equation}
where $\hat{Q}_{in}^{0}$ and $\hat{C}_{in}^{0}$ represent the output proposal query like Eq.\eqref{eq:query_init} and proposal box center of the proposal network respectively. 

\subsection{Label Assignment and Losses}
Following \cite{carion2020end}, we solve the dichotomous matching problem of prediction result and ground truth by the Hungarian algorithm. We define matching cost as the weighted sum of classification cost, regression cost, and 3D IoU cost:
\begin{equation}
Cost=\xi_{1} L_{c l s}(p, \hat{p})+\xi_{2} L_{r e g}(b, \hat{b})+\xi_{3} L_{I o U}(b, \hat{b})
  \label{eq:cost}
\end{equation}
where $L_{cls}$ is the focal loss \cite{lin2017focal}, $L_{reg}$ is the L1 loss between the predicted BEV centers, scale, angle, velocity, and the ground truth, and $L_{IoU}$ is the IoU loss \cite{zhou2019iou} between the predicted boxes and ground truth boxes. $\xi_{1}$, $\xi_{2}$, $\xi_{3}$ are the coefficients of the individual cost terms. Then, we compute both the regression loss and the classification loss given the bipartite matching. 

\section{Implementation Details} \label{sec:imp}
Our implementation is based on the open-source code MMDetection3D \cite{mmdet3d2020}. We use VoxelNet\cite{zhou2018voxelnet} as the 3D backbone and set voxel size to (0.075m, 0.075m, 0.2m) and the detection range to $[ -54m, 54m ]$ for $\mathbb{X}$ and $\mathbb{Y}$ axis and $[ -5m, 3m ]$ for $\mathbb{Z}$ axis. For the image backbone, we use ResNet-50 \cite{he2016deep} and Dual-Swin-Tiny\cite{liang2022cbnet}, both initialized from the instance segmentation model Mask R-CNN \cite{he2017mask} pre-trained on nuImage \cite{caesar2020nuscenes}. We set the full-resolution image size to 896 × 1600. To be friendly with memory and computational consumption, we freeze the weights of the image backbone during training. We set the $N = 400$, $Z = 25$, $M = 3$, $K_{img} = 6$ and $K_{LiD} = 1$ for training and testing. In the next experiments, we use ResNet-50 as the 2D backbone by default to align the baseline.

The data augmentation strategies and training schedules are the same as TransFusion \cite{bai2022transfusion}. we use random rotation with a range of $\theta \in [-\pi/4, \pi/4]$, random box scaling with a factor of $f \in [0.9, 1.1]$, random horizontal flipping in point cloud and images. furthermore, randomly reducing the resolution of LiDAR is set during training. Our training is divided into two stages: Transfusion-L\cite{bai2022transfusion} is chosen as our proposal network, and we first train the LiDAR stream for 20 epochs. Secondly, we keep its weights frozen in the second stage and train the cross-decoder block for 12 epochs with the batch size of 16 using 8 NVIDIA Tesla A100 GPUs. We use Adam optimizer with a one-cycle learning rate schedule, with base learning rate $1 \times 10^{-4}$, weight decay 0.01, and momentum 0.85 to 0.95. It is worth noting that we did not use the depth estimation model during training and did not use Test-Time Augmentation (TTA) or model assembly during inference.

\begin{table*}[t]
  \centering
  \resizebox{2.1\columnwidth}{!}{
  \begin{tabular}{l | c | c c | c c c c c c c c c c}
    \toprule
    Method & Mod. & mAP & NDS & Car & Truck & C.V. & Bus & T.L. &  B.R. & M.T. & Bike & Ped. & T.C. \\
    \midrule
    \textbf{FUTR3D} \cite{chen2022futr3d} & LC & 64.2 & 68.0 & 86.3 & 61.5 & 26.0 & 71.9 & 42.1 & 64.4 & 73.6 & 63.3 & 82.6 & 70.1 \\
    \textbf{BEVFusion-SwinT} \cite{liang2022bevfusion} \dag & LC & 67.9 & 71.0 & 88.6 & 65.0 & 28.1 & 75.4 & 41.4 & 72.2 & 76.7 & 65.8 & 88.7 & 76.9 \\
    \textbf{TransFusion} \cite{bai2022transfusion} & LC & 67.3 & 71.2 & 87.6 & 62.0 & 27.4 & 75.7 & 42.8 & \textbf{73.9} & 75.4 & 63.1 & 87.8 & 77.0 \\
    %    BEVFusion (Liu et al. 2022) LC 68.5 71.4
    %    MSMDFusion-T LC % 69.1 71.8 88.5 64.0 29.2 76.2 44.7 70.4 79.1 68.6 89.7 80.1
    \textbf{CrossFusion} & LC & 69.0 & 71.6 & 88.3 & 66.5 & 32.8 & 75.6 & 43.8 & 70.3 & 77.6 & 67.5 & 89.0 & 78.7 \\
    \textbf{CrossFusion-SwinT} & LC & \textbf{69.7}& \textbf{71.9} & \textbf{88.6} & \textbf{67.6} & \textbf{33.5} & \textbf{76.1} & \textbf{44.6} & 71.2 & \textbf{77.9} & \textbf{68.5} & \textbf{89.2} & \textbf{79.2} \\
    \hline
    \hline
    \textbf{TransFusion-L} \cite{bai2022transfusion} & L &65.5& 70.2& 86.2& 56.7& 28.2& 66.3& 58.8& 78.2& 68.3& 44.2 &86.1& 82.0 \\
    % \textbf{LargeKernel} \cite{chen2022scaling} & L &65.3 & 70.5& 85.9& 55.3& 26.8& 66.2& 60.2& 74.3& 72.5& 46.6 &85.6& 80.0 \\
    \textbf{PointAugmenting} \cite{wang2021pointaugmenting} & LC & 66.8 & 71.0 & 87.5 & 57.3 & 28.0 & 65.2 & 60.7 & 72.6 & 74.3 & 50.9 & 87.9 & 83.6 \\
    \textbf{FusionPainting} \cite{xu2021fusionpainting} & LC & 68.1 & 71.6 & 87.1 & 60.8 & 30.0 & 68.5 & 61.7 & 71.8 & 74.7 & 53.5 & 88.3 & 85.0 \\
    \textbf{TransFusion} \cite{bai2022transfusion} & LC & 68.9 & 71.7 & 87.1 & 60.0 & 33.1 & 68.3 & 60.8 & 78.1 & 73.6 & 52.9 & 88.4 & \textbf{86.7} \\
    \textbf{BEVFusion-SwinT} \cite{liang2022bevfusion} \dag & LC & 69.2 & 71.8 & \textbf{88.1} & 60.9 & 34.4 & 69.3 & \textbf{62.1} & \textbf{78.2} & 72.2 & 52.2 & 89.2 & 85.2 \\
    \textbf{CrossFusion} & LC & 69.2 & 71.8 & 87.4 & 60.4 & 33.8 & 70.3 & 61.0 & 74.3 & 74.3 & 56.1 & 89.2 & 85.1 \\
    \textbf{CrossFusion-SwinT} & LC & \textbf{70.1} & \textbf{72.2} & 87.7 & \textbf{62.4} & \textbf{34.7} & \textbf{71.5} & 61.8 & 75.0 & \textbf{75.1} & \textbf{57.5} & \textbf{89.6} & 85.1 \\
    \bottomrule
      \end{tabular}}
  \caption{Comparison with state-of-the-art methods on nuScenes validation (top) and test (bottom) set. Metrics: mAP($\%$), NDS ($\%$), and AP ($\%$) for each category. ‘C.V.’, ‘T.L.’, 'B.R.', ‘M.T.’, ‘Ped.’, and ‘T.C.’ are short for construction vehicle, trailer, barrier, motor, pedestrian, and traffic cone, respectively. ‘L’ and ‘C’ represent LiDAR and the camera, respectively. We highlight the best performances across all methods with \textbf{bold}. \dag means use depth estimation pre-trained network.}
  \label{tab:val and test}
\end{table*}

\section{Experiments} \label{sec:exp}
In this section, we first compare state-of-the-art methods on nuScenes. Then we designed three experiments with LiDAR corruption to show the modal complementarity of CrossFusion at low-quality LiDAR features. Afterward, we conduct extensive ablation experiments to demonstrate the rationality and importance of each key component of our CrossFusion.

\noindent \textbf{nuScenes Dataset}. The nuScenes dataset\cite{caesar2020nuscenes} is a large-scale autonomous-driving dataset with 3D object annotations that contains 1000 driving sequences, with 700, 150, and 150 sequences for training, validation, and testing, respectively. Each sequence is approximately 20-second long but only provides box annotations every ten frames (0.5s). Each frame contains a wealth of sensor information, including point clouds and carefully calibrated six 360-degree horizontal FOV images at the same timestamp. For 3D detection task, the main metrics are mean Average Precision (mAP) and nuScenes detection score (NDS). The mAP defines a match by considering the 2D center distance thresholds of $0.5m, 1m, 2m, 4m $ across ten classes on the ground plane. NDS is a weighted average of mAP and other attribute metrics, including translation(mATE), scale(mASE), orientation(mAOE), velocity(mAVE), and attribute(mAAE).

\subsection{Main Results}
We compare the results of our CrossFusion on the nuScenes validation set and test set with the state-of-the-art methods. As shown in \cref{tab:val and test}, without model ensemble and depth estimation pre-trained network, our approach surpasses all existing LiDAR-camera fusion methods in both the validation set and test set. our model achieves 69.7\% mAP and 71.9\% NDS in the validation set and 70.1\% mAP and 72.2\% NDS in the test set, respectively. The results show that we can exceed the results of TransFusion on both the validation set(+2.4\% mAP) and the test set(+0.3\% mAP). The nuScenes leaderboard shows that methods with depth priors typically carry more powerful multi-modal  modeling capabilities. For a fair comparison, we replace 2D backbone with Dual-Swin-Tiny\cite{liang2022cbnet}, and we find that CrossFusion can outperform the competitor with depth estimation pre-training, BEVFusion. Meanwhile, our performance has improved in most classes, especially in two small categories, motor, and bike, with mAP improvements of 2.9\% and 3.3\%, respectively. We ascribe the performance improvement to further exploration of the multi-modal perception capability via our interleaving fusion strategy.

\subsection{Robustness against Low-quality LiDAR}
We designed three different experiments with missing point clouds to simulate possible LiDAR malfunctions in autonomous driving scenarios. Experiments show that CrossFusion can exhibit strong robustness without re-training for specific scenarios. To avoid overstatement, We choose Transfusion, a well-known robust model on the nuScenes dataset, as our baseline for comparing the robustness of our method.

\noindent \textbf{Randomly LiDAR Beams Data Augmentation}. To improve the robustness of the model, we propose an effective data enhancement strategy. Inspired by FUTR\cite{chen2022futr3d}, we try to train the network using LiDAR with different resolutions. Following FUTR, we simulated two low-resolution LiDAR beams with 16-beam and 4-beam from the original nuScenes 32-beam LiDAR point clouds. In detail, we first convert the point cloud in the Cartesian coordinate system to the spherical coordinate system, then we select the inclination angle in $\left[-30\degree, 10\degree \right]$ as 4-beam LiDAR. The inclination range of the 16-beam LiDAR is $\left[-7.1\degree, -5.8\degree \right] \cup \left[-4.5\degree, -3.2\degree \right] \cup \left[-1.9\degree, -0.6\degree \right] \cup \left[ 0.7\degree, 2.0\degree \right]$. Data augmentation takes effect after every 10 iterations during training, and we randomly select a low-resolution LiDAR as input. In the following, we use \textbf{CrossFusion*} to indicate that the data augmentation is used.

\begin{figure}[t]
  \centering
  \resizebox{1\columnwidth}{!}{
    \includegraphics[width=1.2\linewidth]{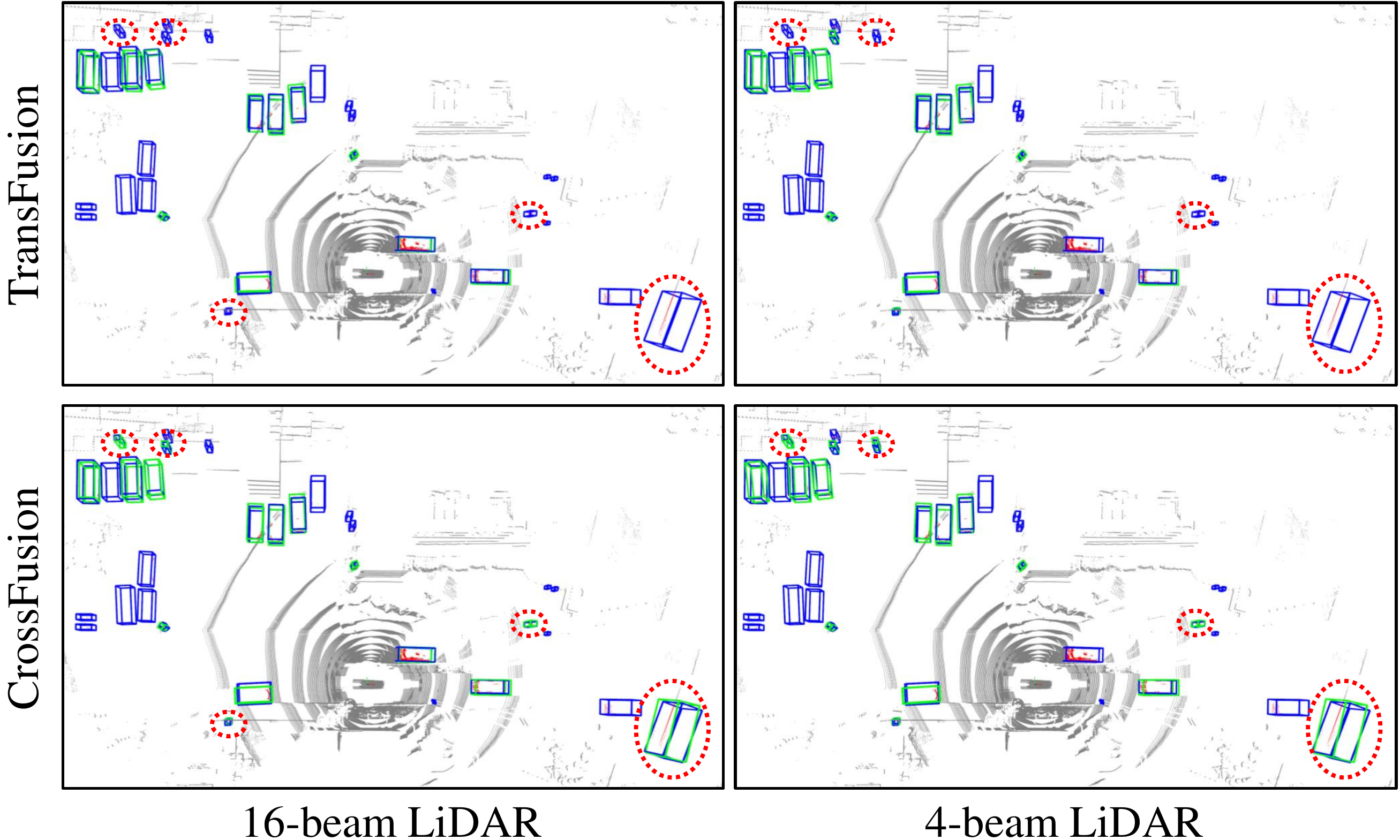}}
   \caption{Results on robustness setting of Low-resolution LiDAR. The red dashed circles represent more recall of our method compared to TransFusion. Blue boxes and green boxes are the ground truth and predictions, respectively. Best viewed with color and zoom-in.}
   \label{fig:beam_visual}
\end{figure}

% \begin{figure}[H]
%   \centering
%   % \fbox{\rule{0pt}{2in} \rule{0.9\linewidth}{0pt}}
%     \includegraphics[width=0.8\linewidth]{fig/block.pdf}
%   \caption{Reduced LiDAR FOV due to truck obscuration in autonomous driving scenarios. The road scene in the red box is shown above. This shows that the obscuration problem is a common problem in autonomous driving scenarios.}
%   \label{fig:block}
% \end{figure}
\noindent \textbf{Low-resolution LiDAR}. LiDAR is highly susceptible to dust or water droplets suspended in the air. For example, in common rainy and foggy weather, LiDAR may not be able to receive the signal reflected by the object due to signal attenuation, which may lead to the problem of reduced resolution. As shown in \cref{tab:beam}, when we used the data augmentation method mentioned above to obtain 16-beam and 4-beam LiDAR point clouds as input, CrossFusion showed a strong ability to resist LiDAR feature degeneration. When the number of beams of LiDAR is reduced to 16, our method improves by 3.8\% mAP and 1.8\% NDS compared to TransFusion. Surprisingly when the input is 4-beam, our model has a 3.6\% mAP and 1.4\% NDS improvement compared to Transfusion, even without data augmentation. If we enable data augmentation, the gain of mAP and NDS will be increased to 5.2\% and 2.4\%, respectively. We speculate that the performance enhancement is because the interleaving fusion strategy helps to exploit modal complementarity. The results reveal that randomly LiDAR beams augmentation can drive the model to mine more connections between image features and LiDAR features for better robustness. As can be seen from the visualization in \cref{fig:beam_visual}, we can obtain more recall with low-resolution LiDAR input, especially in the challenging pedestrian category, which may better guarantee the safety of autonomous driving. In addition, low-resolution LiDAR is usually more affordable, and our model exhibits the possibility of deploying at a low cost.

\begin{figure}[t]
  \centering
    \includegraphics[width=1\linewidth]{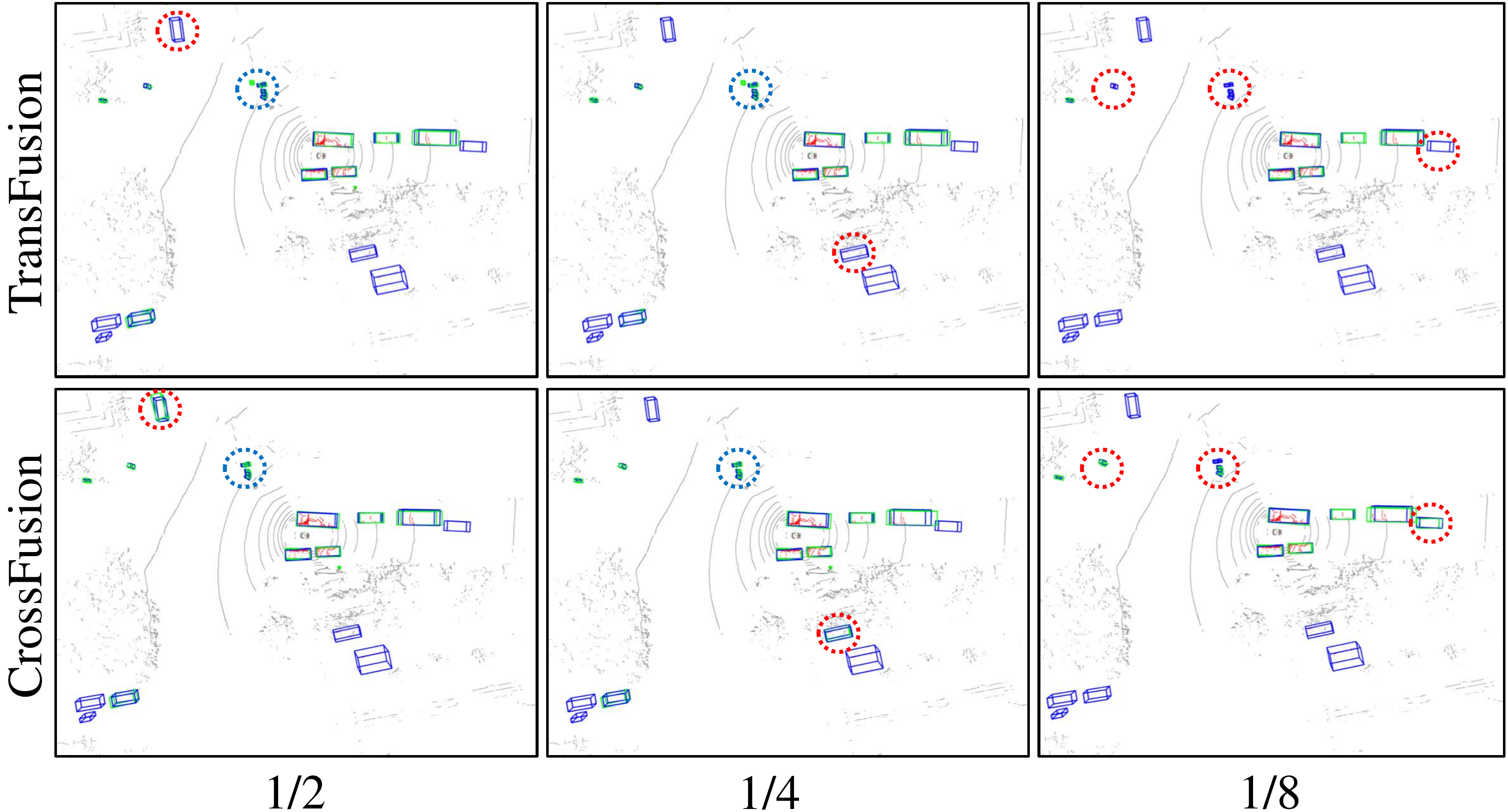}

   \caption{Results on robustness setting of reduced points. The red dashed circles represent more recall of our method compared to TransFusion, and the blue dashed circles represent error detection. Blue boxes and green boxes are the ground truth and predictions, respectively. Best viewed with color and zoom-in.}
   \label{fig:reduce_point}
\end{figure}

\noindent \textbf{Reduced Number Of Points}. LiDAR is susceptible to surrounding light and vulnerable to the reflection rate of objects. The point cloud is easily lost in outdoor environments when exposed to bright light or highly reflective materials. \cref{tab:drop points} presents the results of reducing the points to half, a quarter, and an eighth of the original points, respectively. It is proof that CrossFusion has better results for all reduction ratios, and the advantage of our method becomes more obvious the larger the reduction ratio is. Noteworthy, when the points are only one-eighth of the original points, CrossFusion can increase mAP and NDS by 4.6\% and 2.5\%, respectively, compared to Transfusion. This demonstrates that CrossFusion is less dependent on LiDAR features and has the potential for modal complementarity. As shown in the \cref{fig:reduce_point}, our method can compensate for the deficiencies of LiDAR target point clouds to a certain extent. When reduced to a quarter of the point cloud, as shown in the blue dashed circles, Transfusion has false detections due to missing point cloud features, while our model can still show satisfactory detection results.

\begin{table}[!ht]
    % 线数鲁棒
  \centering
  \resizebox{0.9\columnwidth}{!}{
  \begin{tabular}{c | c | c c}
    \toprule
    Method & Beam & mAP & NDS \\
    \midrule
    % TransFusion \cite{bai2022transfusion} & \multirow{3}{*}{32} & 67.5 & 71.3 \\
    % CrossFusion & & 69.0(+1.5) & 71.6(+0.3) \\
    % CrossFusion* & & \textbf{69.0(+1.5)} & \textbf{71.6(+0.3)} \\
    % \hline
    TransFusion \cite{bai2022transfusion} & \multirow{3}{*}{16} & 51.5 & 61.8 \\
    CrossFusion & & 54.8(+3.3) & 63.3(+1.5) \\
    CrossFusion* & & \textbf{55.3(+3.8)} & \textbf{63.6(+1.8)} \\
    \hline
    TransFusion \cite{bai2022transfusion} & \multirow{3}{*}{4} & 39.2 & 54.3 \\
    CrossFusion & & 42.8(+3.6) & 55.9(+1.4) \\
    CrossFusion* & & \textbf{44.4(+5.2)} & \textbf{56.7(+2.4)} \\
    \bottomrule
  \end{tabular}}
  \caption{Low-resolution LiDAR simulated on the nuScenes validation dataset as input for mAP and NDS. Beam settings are the same as Randomly LiDAR Beams Data Augmentation. The numbers in parentheses represent the improvement compared to the baseline.}
  \label{tab:beam}
\end{table}

\begin{table}[ht]
    % 随机删除点
  \centering
  \resizebox{0.85\columnwidth}{!}{
  \begin{tabular}{c | c | c c}
    \toprule
    Method & Ratio & mAP & NDS \\
    \midrule
    TransFusion \cite{bai2022transfusion} & \multirow{3}{*}{1/2} & 63.8 & 68.5 \\
    CrossFusion & & 66.1+(2.3) & 69.6+(0.3) \\
    CrossFusion* & & \textbf{66.1(+2.3)} & \textbf{69.6(+0.3)} \\
    \hline
    TransFusion \cite{bai2022transfusion} & \multirow{3}{*}{1/4} & 56.4 & 63.7 \\
    CrossFusion &  & 59.3(+1.9) & 65.2(+1.5) \\
    CrossFusion* & & \textbf{59.7(+2.3)} & \textbf{65.4(+1.7)} \\
    \hline
    TransFusion \cite{bai2022transfusion} & \multirow{3}{*}{1/8} & 44.6 & 55.8 \\
    CrossFusion & & 48.4(+3.8) & 57.8(+2.0) \\
    CrossFusion* & & \textbf{49.2(+4.6)} & \textbf{58.3(+2.5)} \\
    \bottomrule
  \end{tabular}}
  \caption{Random reduction of points results in the nuScenes validation dataset. “Ratio” indicates the ratio of the input points to the original points after random reduction.}
  \label{tab:drop points}
\end{table}

\noindent \textbf{Reduced LiDAR Field Of View}. The most vulnerable part of the spinning LiDAR is the rotating mechanism, and it will bring the problem of missing Field of view (FOV). To simulate the above situation, we adjusted the FOV of the LiDAR, and the results are shown in \cref{tab:drop view}. Experimental results show that our method has mAP and NDS gain at each set compared to the baseline. This demonstrates that CrossFusion is more robust, and our fusion method has a certain compensation effect on the reduction of LiDAR FOV.
\begin{table}[ht]
    % 视野缺失
  \centering
  \resizebox{0.88\columnwidth}{!}{
  \begin{tabular}{c | c | c c}
    \toprule
    Method & FOV & mAP & NDS \\
    \midrule
    % TransFusion \cite{bai2022transfusion} & \multirow{3}{*}{$-\pi, \pi$} & 67.5 & 71.3 \\
    % CrossFusion & & 69.0(+1.5) & 71.6(+0.3) \\
    % CrossFusion* & & \textbf{69.0(+1.5)} & \textbf{71.6(+0.3)} \\
    % \hline
    TransFusion \cite{bai2022transfusion} & \multirow{3}{*}{$-\frac{5 \pi}{6},\frac{5 \pi}{6}$} & 50.1 & 62.2 \\
    CrossFusion & & 51.6(+1.5) & 62.8(+0.6) \\
    CrossFusion* & & \textbf{51.8+(1.7)} & \textbf{62.9(+0.7)} \\
    \hline
    TransFusion \cite{bai2022transfusion} & \multirow{3}{*}{$-\frac{2 \pi}{3},\frac{2 \pi}{3}$} & 38.7 & 56.4 \\
    CrossFusion &  & 40.0(+1.3) & 56.9(+0.5) \\
    CrossFusion* & & \textbf{40.1(+1.4)} & \textbf{56.9(+0.5)} \\
    \hline
    TransFusion \cite{bai2022transfusion} & \multirow{3}{*}{$-\frac{\pi}{2},\frac{\pi}{2}$} & 29.6 & 51.6 \\
    CrossFusion & & 30.6(+1.0) & 52.0(+0.4) \\
    CrossFusion* & & \textbf{30.7(+1.1)} & \textbf{52.0(+0.4)} \\
    \hline
    TransFusion \cite{bai2022transfusion} & \multirow{3}{*}{$-\frac{\pi}{3},\frac{\pi}{3}$} & 20.5 & 46.4 \\
    CrossFusion & & 21.1(+0.6) & 46.5(+0.1) \\
    CrossFusion* & & \textbf{21.2(+0.7)} & \textbf{46.6(+0.2)} \\
    \bottomrule
  \end{tabular}}
  \caption{Compare the results of CrossFusion at different FOVs on the nuScenes validation set. Numbers in parentheses represent an improvement over the baseline.}
  \label{tab:drop view}
\end{table}

\begin{table}[ht]
    % 这里是8e的结果，一层encoder和3层block的结果
  \centering
  \resizebox{0.72\columnwidth}{!}{
  \begin{tabular}{c | c c c | c c}
    \toprule
    Idx & CDB & IFE & PSE & mAP & NDS \\
    \midrule
    a) & - & -  & - & 65.2 & 69.9 \\
    b) & $\surd$ & - & - & 67.9 & 71.1 \\
    c) & $\surd$ & $\surd$ & - &  67.9 & 71.2 \\  
    d) & $\surd$ & - & $\surd$ &  68.0 & 70.9 \\  
    e) & $\surd$ & $\surd$ & $\surd$ &  \textbf{68.5} & \textbf{71.3} \\  

    \bottomrule
  \end{tabular}}
  \caption{Ablation of our key component. ``CDB" and ``IFE"  means Cross Decoder Block and Image Feature Encoder, respectively, and ``PSE" represents Point Select Enhancement.}
  \label{tab:Ablation}
\end{table}

\subsection{Ablation Studies}
To demonstrate the effectiveness of each component of CrossFusion, we performed an ablation study of them on the nuScenes validation set. For fast iteration, we train 8 epochs by default.

\noindent \textbf{Ablation of key components}. As shown in \cref{tab:Ablation}, the first line indicates the result of the proposal network. \textbf{b)} When the cross-decoder block is added, mAP and NDS can be improved by 2.7\% and 1.2\%, respectively. This shows that the strategy of using interleaving fusion, which is able to interact with image features and BEV features, provides richer instance-level features, gaining 2.7\% mAP and 1.2\% NDS. \textbf{c)} NDS 0.1\% gain when the image feature encoder is added separately. This is because Cross decoder blcok has fully interacted with the instance features and point cloud features of the image. Adding more image features will only continue to optimize the bounding box that is already close to the ground truth. \textbf{d)} When point select enhancement is added, the network will be more sensitive to learning difficult samples, bringing 0.1\% mAP gain and leading to the loss of bounding box accuracy. \textbf{e)} With the combination of the above two modules, the mAP is improved by 0.6\% compared to \textbf{b)}. This shows that with the help of point select enhancement, higher-quality image information needs to be introduced to effectively help learn hard samples.

\noindent \textbf{Number Of Fusion Components}. As shown in \cref{tab:encoder layers}, a two-layer image feature encoder is appropriate. Adding more encoders may lead to overfitting and thus affect performance. In \cref{tab:decoder layers}, increasing the number of cross-decoder blocks to three allows for continuous performance growth. So we choose this parameter in our model to achieve the best results.

\begin{table}[ht]
    % IFE层数的消融
  \centering
  \resizebox{0.84\columnwidth}{!}{
  \begin{tabular}{c | c c | c c c}
    \toprule
    $L_{IFE}$ & mAP & NDS & mATE & mASE & mAOE \\
    \midrule
    1 & 68.5 & 71.3 & \textbf{28.3} & 25.8 & 29.6 \\
    2 & \textbf{68.8} & \textbf{71.4} & 28.5 & \textbf{25.8} & 29.2 \\  
    3 & 68.3 & 71.1 & 28.4 & 26.2 & \textbf{28.9} \\
    \bottomrule
  \end{tabular}}
  \caption{Performance among different encoder layers, $L_{IFE}$ represents the number of layers of the image feature encoder.}
  \label{tab:encoder layers}
\end{table}

\begin{table}[ht]
    % CDB层数的消融
  \centering
  \resizebox{0.84\columnwidth}{!}{
  \begin{tabular}{c | c c | c c c}
    \toprule
    $L_{CDB}$ & mAP & NDS & mATE & mASE & mAOE\\
    \midrule
    2 & 68.2 & 71.2 & \textbf{28.4} & \textbf{25.6} & 29.2 \\
    3 & \textbf{68.8} & \textbf{71.4} & 28.5 & 25.8 & \textbf{29.2} \\  
    4 & 68.4 & 71.0 & 28.6 & 26.4 & 29.5 \\
    \bottomrule
  \end{tabular}}
  \caption{The impact of the number of cross decoder blocks on performance, $L_{CDB}$ represents the number of cross decoder blocks.}
  \label{tab:decoder layers}
\end{table}

\begin{table}[ht]
    % order of decoders
  \centering
  \resizebox{0.91\columnwidth}{!}{
  \begin{tabular}{c| c | c c | c c c}
    \toprule
    Idx & Order & mAP & NDS & mATE & mASE & mAOE\\
    \midrule
    a) & (LC)3 & 65.5 & 69.2 & 28.9 & \textbf{25.3} & 30.9 \\
    b) & 3L3C & 65.5 & 69.4 & 28.9 & 25.7 & 30.1 \\
    c) & 3C3L & 67.8 & 71.0 & 28.4 & 25.8 & \textbf{28.8} \\
    d) & \textbf{(CL)3} & \textbf{68.5} & \textbf{71.3} & \textbf{28.3} & 25.8 & 29.6 \\
    \bottomrule
  \end{tabular}}
  \caption{The ablation of the decoding order inside the cross-decoder block.}
  \label{tab:order of decoders}
\end{table}
% \vspace{-1em}

\noindent \textbf{The Order Of Decoding}. We designed four different decoding orders to explore the impact of different decoder arrangements on CrossFusion performance. 
% We choose the order of our blocks as the default setting. 
For the convenience of representation, we will write the LiDAR decoder as ``L" and the image decoder as ``C". Thus our setting can be denoted by ``(CL) 3", which means 3-layer combination consisting of first ``C" and then ``L". Analogously, ``3L3C'' means in series of first 3 layers of ``L" then 3 layers of ``C".  Below are other combinations we select, which are (LC)3 and 3L3C.
% means in series 3 layers of ``L" and 3 layers of ``C". Similarly ``(CL)" 3 and ``3C3L swap the positions of ``L" and ``C". 
The \cref{tab:order of decoders} shows that the performance of both \textbf{a)} and \textbf{b)} is lower than that of \textbf{c)} and \textbf{d)}. We speculate it is not reasonable to try to update the LiDAR proposal with accurate 3D information using image features without depth information. Under the premise of determining a reasonable order, \textbf{d)},\textbf{c)} the interleaving fusion strategy outperforms the serial fusion approach by 0.7\% mAP. The experimental results show that the decoder architecture with interleaving fusion can make the model have stronger multi-modal modeling capability.

\section{Conclusion}
In this paper, we propose CrossFusion, a novel LiDAR-camera fusion framework without depth pre-training. We design the interleaving fusion strategies with multi-level proposal query enhancements to pursue equilibrium and complementarity between modalities. Extensive experiments have demonstrated the robustness and noise-resistance of our model, which better mines image information to combat LiDAR corruption. We hope that our work will stimulate further exploration of robust multi-modal fusion in autonomous driving scenarios.

%%%%%%%%% REFERENCES
{\small
\bibliographystyle{ieee_fullname}
\bibliography{CrossFusionbib}
}

\end{document}